\documentclass[sigconf]{acmart}

\AtBeginDocument{%
  }

\setcopyright{acmlicensed}
\copyrightyear{2026}
\acmYear{2026}
\acmDOI{}
\acmConference[BioKDD '26]{ACM SIGKDD Workshop on Data Mining in Bioinformatics}{August 2026}{TBD}
\acmISBN{}

\usepackage{tabularx}
\usepackage{booktabs}
\usepackage{makecell}
\usepackage{placeins}

\begin{document}

%% ── Title ────────────────────────────────────────────────────────────────────
\title{SCTA: An Agentic Framework for Stable and Interpretable Target Gene Discovery from Single-Cell RNA Sequencing}

%% ── Authors ──────────────────────────────────────────────────────────────────
\author{Shuyu Chen}
\affiliation{%
\department{Siebel School of Computing and Data Science}
  \institution{University of Illinois Urbana-Champaign}
  \city{Champaign}
  \state{IL}
  \country{USA}}

\author{Chen Zhu}
\affiliation{%
\department{School of Information Sciences}
  \institution{University of Illinois Urbana-Champaign}
  \city{Champaign}
  \state{IL}
  \country{USA}}

\author{Ye Zhang}
\affiliation{%
  \institution{Novartis Biomedical Research}
  \department{Biomedical Research}
  \city{Cambridge}
  \state{MA}
  \country{USA}}

\author{Yang Li}
\orcid{0000-0003-2530-7167}
\affiliation{%
\department{Department of Ecology and Evolutionary Biology}
  \institution{University of Michigan}
  \city{Ann Arbor}
  \state{MI}
  \country{USA}}

\author{Qiqi Xie}
\affiliation{%
\department{School of Information Sciences}
  \institution{University of Illinois Urbana-Champaign}
  \city{Champaign}
  \state{IL}
  \country{USA}}

\author{Haohan Wang}
\orcid{0000-0002-1826-4069}
\email{haohanw@illinois.edu}
\affiliation{%
\department{School of Information Sciences}
  \institution{University of Illinois Urbana-Champaign}
  \city{Champaign}
  \state{IL}
  \country{USA}}

\renewcommand{\shortauthors}{Chen et al.}

%% ── Abstract ─────────────────────────────────────────────────────────────────
\begin{abstract}
Identifying therapeutic target genes from single-cell RNA sequencing (scRNA-seq) data remains a fundamental challenge in translational biology. Unlike bulk assays, scRNA-seq captures heterogeneous cellular states and rare subpopulations, but this same heterogeneity makes target discovery highly sensitive to analytical choices throughout the pipeline, including preprocessing, cell population selection, differential expression analysis, and downstream biological interpretation. As a result, existing workflows and general-purpose analysis agents often produce unstable or difficult-to-interpret target hypotheses, limiting their reliability for disease-focused discovery. We present \textbf{SCTA} (Single-Cell Target Agent), a decision-centric agentic framework for stable and interpretable target gene discovery from scRNA-seq data. Rather than treating analysis as a single general-purpose reasoning task, SCTA decomposes target discovery into specialized agents aligned with key decision points in the single-cell pipeline and constrains downstream reasoning with structured biological evidence. In a representative ablation study on hereditary chronic pancreatitis, we demonstrate that SCTA's full evidence integration yields the most stable target selection across independent runs among the tested configurations, while recovering biologically coherent, disease-relevant mechanisms validated in prior studies. These results suggest that decision-aware agent orchestration tailored to the structure of single-cell analysis can improve the robustness, interpretability, and practical utility of target discovery in precision medicine.
\end{abstract}

%% ── CCS Concepts ─────────────────────────────────────────────────────────────
%% Generated from https://dl.acm.org/ccs/ccs.cfm



\maketitle

%% ── Introduction ─────────────────────────────────────────────────────────────
\section{Introduction}
Single-cell RNA sequencing (scRNA-seq) is widely used to identify disease-relevant cell populations and candidate therapeutic targets. In practice, however, target discovery from single-cell data is fragile: small changes in analytical choices can produce different target lists, alter downstream validation priorities, and reduce reproducibility across cohorts \cite{freedman2015reproducibility}.

A central source of this fragility is structured biological heterogeneity. scRNA-seq captures mixtures of discrete cell types, continuous states, and rare subpopulations. Under these conditions, core analysis steps can diverge across methods, including clustering and trajectory inference \cite{duo2018clustering, saelens2019trajectory}. Upstream choices in normalization, integration, clustering, and annotation therefore reshape the downstream candidate space rather than simply denoising fixed signals \cite{ge2025normalization, nguyen2023integration, dann2022da}.

This is especially problematic for translational discovery because many perturbations do not trigger explicit failure signals: pipelines can remain internally coherent while yielding materially different targets. Meta-analytic evidence also shows that reproducibility of single-study differential expression can be limited, while consistency-aware prioritization improves reliability \cite{nakatsuka2025meta}. For target discovery, robustness to plausible analytical perturbation is therefore a core requirement.

General-purpose agent frameworks automate analysis and tool use, but they usually do not make single-cell-specific decision points auditable or explicitly optimize for stable target convergence across runs \cite{hong2024metagpt, hong2024datainterpreter, huang2025biomni}. We address this gap with \textbf{S}ingle-\textbf{C}ell \textbf{T}arget \textbf{A}gent (\textbf{SCTA}), a decision-centric multi-agent framework that decomposes target discovery into specialized, constrained stages. SCTA makes cell-type triage, candidate filtering, and evidence aggregation explicit, enabling target prioritization to be evaluated by cross-run stability rather than single-run plausibility.

\section{Related Work}
 
\subsection{Single-cell analysis pipelines and analytical instability}
 
Established single-cell toolkits such as Seurat, Scanpy, and Monocle
provide modular workflows for normalization, clustering, trajectory
inference, and annotation~\cite{seurat2015spatial,wolf2018scanpy,trapnell2014monocle},
and deep generative approaches such as scVI and scANVI improve
representation learning and annotation under batch effects or
incomplete labels~\cite{lopez2018scvi,xu2021scanvi}. Best-practice
syntheses consolidate these components into recommended pipelines and
emphasize that no single configuration is universally
optimal~\cite{luecken2019best,heumos2023best}. Systematic benchmarks
further show that pipeline choice materially affects downstream
results: clustering, integration, trajectory inference, and
differential-abundance methods can disagree substantially on the same
input data~\cite{duo2018clustering,saelens2019trajectory,luecken2022benchmarking,tian2019benchmarking,nguyen2023integration,ge2025normalization,dann2022da}.
 
A complementary line of work argues that single-cell differential
expression is itself a major source of analytical
fragility. \citet{squair2021confronting} show that single-cell DE
tests can produce inflated false-discovery rates when pseudo-replication
is ignored, and meta-analytic evidence indicates that single-study DE
results are often unstable, while consistency-aware prioritization
improves reliability~\cite{nakatsuka2025meta}. Combined with the
broader reproducibility concerns documented in preclinical
research~\cite{freedman2015reproducibility}, these observations imply
that upstream choices in normalization, integration, clustering, and
annotation reshape the downstream candidate space rather than simply
denoising a fixed signal~\cite{dann2022da,ge2025normalization,nguyen2023integration}.
For translational target discovery, the relevant failure mode is
therefore not only inaccurate intermediate labels, but
\emph{decision sensitivity of the final target set itself}—a property
that existing evaluations, which tend to emphasize annotation accuracy,
integration quality, or runtime, do not directly measure.
 
\subsection{LLM agents and tool-augmented reasoning}
 
Large language models have rapidly evolved from single-turn
predictors into agents that plan, call external tools, and iterate
over intermediate results. Foundational techniques such as
chain-of-thought prompting~\cite{wei2022chain} and
self-consistency~\cite{wang2022selfconsistency} improved multi-step
reasoning, while ReAct~\cite{yao2023react} introduced interleaved
reasoning and acting, and Reflexion~\cite{shinn2023reflexion} added
verbal self-feedback for iterative refinement. In parallel,
tool-learning frameworks such as Toolformer~\cite{schick2023toolformer}
demonstrated that models can be trained to invoke external APIs, and
multi-agent orchestration systems such as
AutoGen~\cite{wu2023autogen}, MetaGPT~\cite{hong2024metagpt}, and
Data Interpreter~\cite{hong2024datainterpreter} extended this paradigm
to collaborative task decomposition and code execution.
 
These systems substantially expand what LLMs can accomplish, but they
are typically optimized for general-purpose task completion or code
generation rather than for stability of a downstream scientific
decision. In particular, they do not formalize domain-specific
decision points, and their evaluations rarely measure repeated-run
convergence of the final outputs—an attribute that becomes critical
when the downstream cost of an unreliable recommendation is high. SCTA
inherits the multi-agent and tool-use abstractions from this
literature but constrains agent actions to a fixed, biologically
motivated decision pipeline and evaluates the resulting target set
under repeated execution.
 
\subsection{Agentic systems for biomedical and scientific discovery}
 
A growing body of work applies LLMs and LLM-driven agents to
biomedical and scientific
problems~\cite{balaji2023gptmolberta,lin2023evolutionary,madani2023progen,qi2024hypothesis}.
General-purpose scientific agents such as
ChemCrow~\cite{bran2023chemcrow} and
Coscientist~\cite{boiko2023autonomous} demonstrate that tool-augmented
LLMs can plan and execute non-trivial chemistry workflows, while
biomedical foundation models such as
BioGPT~\cite{luo2022biogpt} and Med-PaLM~\cite{singhal2023medpalm}
show strong performance on clinical and literature-grounded reasoning
tasks. In single-cell and broader biomedical analysis, systems such
as CellTypeAgent~\cite{chen2025celltypeagent},
scAgent~\cite{mao2025scagent}, CellAgent~\cite{xiao2024cellagent},
CellVoyager~\cite{alber2025cellvoyager}, and Biomni~\cite{huang2025biomni}
emphasize annotation quality, end-to-end pipeline automation, or
broad exploratory capability.
 
Although several of these systems can execute single-cell pipelines,
they differ from SCTA in optimization objective. CellTypeAgent and
scAgent treat annotation as the primary
goal~\cite{chen2025celltypeagent,mao2025scagent}.
CellAgent automates end-to-end pipeline
execution~\cite{xiao2024cellagent}. CellVoyager targets autonomous
exploration and hypothesis diversity~\cite{alber2025cellvoyager}.
Biomni provides broad cross-domain biomedical task
execution~\cite{huang2025biomni}. None of these systems explicitly
formalizes decision-critical stages in single-cell analysis or
evaluates target-gene stability under repeated analytical runs—the
central concern for translational discovery. This distinction matters
because high annotation quality or smooth automation does not
guarantee convergence of the final target shortlist. Our focus is
different: SCTA is designed around decision-centric orchestration for
stable target discovery, with evaluation centered on repeated-run
convergence of the final target set rather than single-run
plausibility alone.

\section{Methods}

\begin{figure*}[t]
\centering
\includegraphics[width=0.9\linewidth]{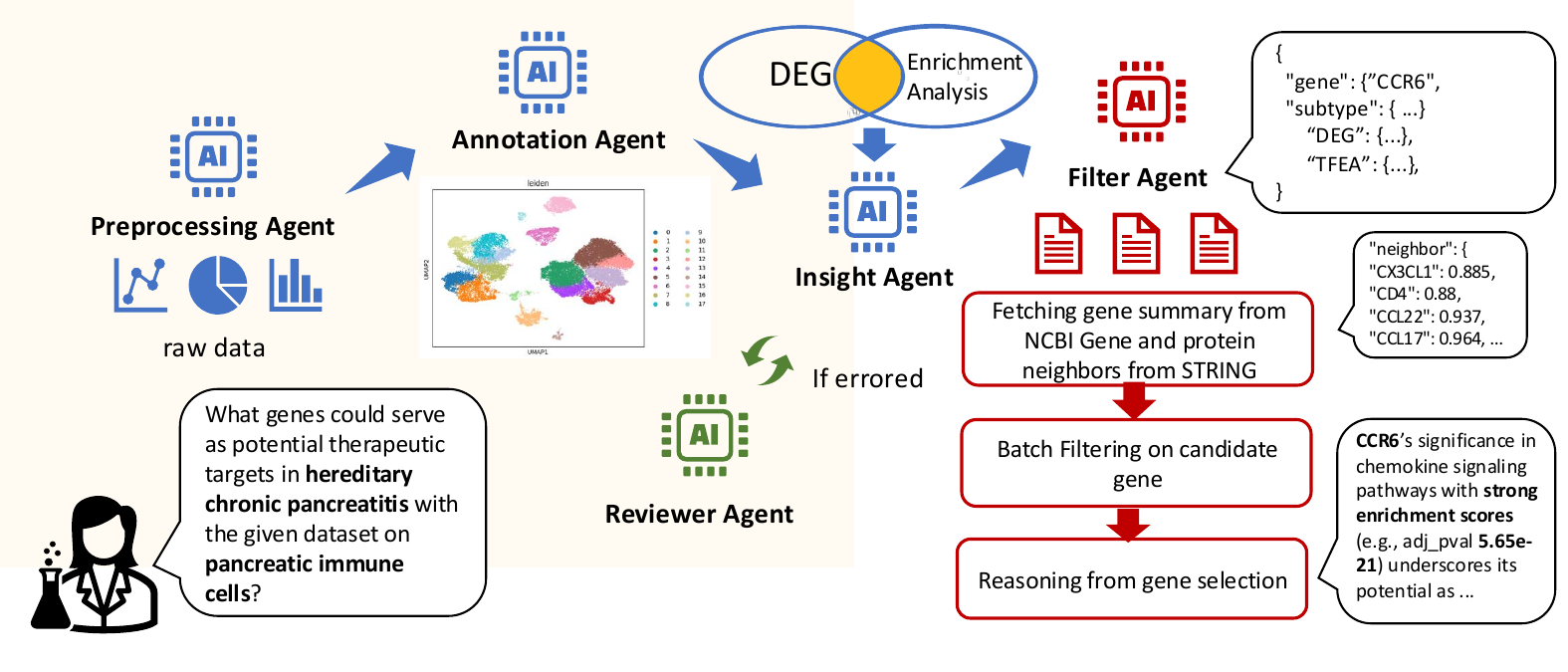}
\caption{SCTA framework architecture. SCTA structures single-cell target discovery around explicit, biologically constrained decision points. Procedural analyses generate candidate signals, which are progressively narrowed through enrichment-guided selection and gene-level reasoning grounded in external biological evidence; a failure-handling agent supports controlled repair when execution errors occur. This design makes how analytical choices shape final target prioritization transparent, interpretable, and more stable under heterogeneity.}
\label{fig:pipeline}
\Description{Architecture diagram showing the SCTA multi-agent framework with five specialized agents (Preprocessing, Annotation, Insight, Filter, and Reviewer) arranged in a sequential pipeline. Each agent has a dedicated tool box and produces structured outputs that flow unidirectionally to downstream stages. The diagram illustrates how raw single-cell data is progressively refined through cell-type annotation, disease-context selection, enrichment-guided filtering, and evidence aggregation to produce a final prioritized target gene set.}
\end{figure*}

\subsection{Agent Specialization and Roles}
SCTA decomposes standard single-cell analysis into specialized agents with fixed execution order and structured handoffs (Figure~\ref{fig:pipeline}). This separation localizes high-impact decisions and improves auditability of how upstream choices affect final targets.

\textbf{Preprocessing Agent.}
Performs deterministic preprocessing, including quality control, normalization, feature selection, dimensionality reduction, clustering, and cluster-level differential expression. It outputs ranked differentially expressed genes (DEGs) with statistics.

\textbf{Annotation Agent.}
Assigns biological identities to clusters using DEG signatures, enrichment patterns, and prior knowledge, producing an explicit cluster-to-cell-type mapping used by downstream stages.

\textbf{Insight Agent.}
Selects disease-informative cell populations and forms a candidate gene pool by combining differential expression with enrichment-level evidence.

\textbf{Filter Agent.}
Aggregates multiple evidence streams to produce a fixed-size prioritized target set, including DEG strength, subtype context, enrichment support, interaction context, and concise functional summaries.

\textbf{Reviewer Agent.}
Triggered only when execution errors occur; reviews failed steps and supports controlled repair under retry limits without changing the biological objective.

\subsection{Architecture}
The full implementation is available at \url{https://github.com/silviachen46/SCTA}. SCTA follows three principles: localized decision-making, constrained agent actions, and controlled error propagation.

\textbf{Tool Box abstraction.}
Each agent receives a disjoint, predefined tool set and cannot access tools outside scope or define new tools at runtime. This reduces off-path execution and keeps each stage within intended analytical boundaries.

\textbf{Unidirectional information flow.}
Outputs propagate sequentially from preprocessing to annotation, insight, and filtering. This enforces explicit conditioning of downstream prioritization on upstream decisions.

\textbf{Failure handling.}
We define a binary failure indicator $I_{\text{fail}}(x)\in\{0,1\}$ over agent outcomes. When failure is detected, control transfers to the Reviewer Agent. The loop terminates on success or after retry budget exhaustion, enabling graceful abort with explicit failure state.

\subsection{Ablation Experimental Design}
To characterize how biological evidence modules affect downstream target-selection stability, we conducted a focused ablation study on hereditary chronic pancreatitis (Pancreatitis-Her). This disease context was selected as a mechanistic case study because it exhibits a well-characterized inflammatory immune program and supports direct comparison of how evidence perturbations reshape recurrent target selection.

To isolate evidence-module effects from upstream preprocessing variability, all ablation settings reused fixed Pancreatitis preprocessing artifacts, including the preprocessed AnnData object, cell-type annotation outputs, and intermediate graph results. We then repeated only the downstream target-selection stages of SCTA for six independent runs per setting. This design explicitly evaluates \emph{downstream target-selection stability} under fixed preprocessing artifacts; it does not measure full end-to-end preprocessing stability.

Three settings were evaluated:

\textbf{Normal.} The full SCTA downstream evidence configuration.

\textbf{BioKnowledge\_ablation.} STRING neighbor retrieval and gene-summary evidence removed, eliminating network-context and concise functional-knowledge inputs during filtering.

\textbf{EnrichmentFunction\_ablation.} TF/pathway enrichment evidence removed from the Insight stage, eliminating regulatory and pathway-level support before final filtering.

Within-setting stability is quantified by the average pairwise Jaccard similarity across the six independent runs. We also summarize recurrent targets using stable genes selected in at least $3/6$ runs, with stricter summaries at $4/6$ and $6/6$. This protocol allows us to test whether SCTA's full evidence integration improves the reproducibility and biological coherence of downstream target prioritization when upstream preprocessing is held fixed.

\section{Experiments and Results}

\subsection{Datasets}
We evaluated SCTA on three publicly available scRNA-seq datasets from GEO, representing oncological, systemic inflammatory, and tissue-resident chronic inflammatory conditions \cite{geo}.

\textbf{GSE193337} contains scRNA-seq profiles of human prostate tumors with matched benign tissues, capturing tumor epithelial cells and the tumor microenvironment \cite{GSE193337}.

\textbf{GSE149689} consists of PBMC scRNA-seq profiles from healthy donors and patients with mild/severe COVID-19 or severe influenza, representing a systemic inflammatory context \cite{GSE149689}.

\textbf{GSE165045} includes CITE-seq and paired scTCR-seq profiles of pancreatic immune cells from hereditary and idiopathic chronic pancreatitis patients and non-diseased controls \cite{GSE165045}.

All sample-level files within each study were merged into a single AnnData object. No explicit batch integration was applied, preserving disease-associated expression differences.

\subsection{Cell-Type Selection and Target Gene Attribution}

\begin{table*}[t]
\centering
\caption{Cell-type selection and representative target gene attribution by SCTA across datasets.}
\label{tab:celltype_attribution}
\begin{tabularx}{\textwidth}{
l
>{\centering\arraybackslash}p{1.6cm}
>{\raggedright\arraybackslash}p{6.2cm}
>{\raggedright\arraybackslash}X
}
\toprule
\textbf{Dataset}
& \textbf{Cell types annotated}
& \textbf{Cell types selected for investigation}
& \textbf{Representative target genes (cell subtype)} \\
\midrule

\makecell{Prostate adenocarcinoma\\(GSE193337)}
& 14
& B cells; Dendritic cells; \textbf{Epithelial cells}; Macrophages; NK cells; T cells
& BNIP3 (Prostatic epithelial cells); CTSL (Transitional epithelial cells); KLK3 (Luminal epithelial cells) \\
\midrule
\makecell{Chronic pancreatitis\\(GSE165045)}
& 11
& \textbf{Dendritic cells}; Macrophages; Mast cells; \textbf{Memory T cells}; Monocytes; NK cells; Stressed cells
& CCL20 (Dendritic cells); CCR6 (T cells); HSP90AA1 (Dendritic cells) \\
\midrule
\makecell{COVID-19\\(GSE149689)}
& 14
& Activated T cells; B cells; Cytotoxic T cells; \textbf{Dendritic cells}; \textbf{Interferon-stimulated cells}; NK cells; T cells
& IL1B (Activated monocytes); CCL3 (Activated monocytes); HLA-DRB5 (Type I interferon-stimulated cells) \\
\bottomrule
\end{tabularx}
\end{table*}

Table~\ref{tab:celltype_attribution} summarizes cell-type selection and representative target gene attribution from one representative run per dataset. SCTA annotates a diverse set of cell types and then selects a disease-relevant subset for downstream analysis. This explicit selection step constrains the analytical space and exposes a decision stage that is typically implicit in automated pipelines. Candidate targets are linked to specific cellular sources, providing cell-resolved interpretability that is obscured in bulk-level approaches.

\subsection{Comparison with Baseline}

\subsubsection{Baseline Design}
We implemented a single-pass LLM baseline using the same preprocessed data and analysis functions as SCTA. In the first stage, an LLM constructs and executes a standard pipeline including quality control, normalization, clustering, cell type annotation with CellTypist \cite{dominguezconde2022hca}, and Wilcoxon differential expression via Scanpy \cite{wolf2018scanpy}. In the second stage, a separate LLM selects $k=10$ target genes from the results. The baseline is executed once per dataset.

\subsubsection{Evaluation Protocol}
All methods use the same LLM backend (GPT-4o \cite{openai2024gpt4o}), identical input data, and identical preprocessing. SCTA and Biomni are each run $R=10$ times per dataset; a consensus set is formed by retaining the top-$k$ genes by selection frequency. We evaluate on five criteria: (1) \textbf{disease relevance} via Open Targets Q3 score \cite{opentargets2021}; (2) \textbf{functional coherence} via number of significantly enriched Reactome\_2022 pathways and mean enrichment score \cite{gillespie2022reactome,liberzon2011msigdb}; (3) \textbf{runtime}; and (4) \textbf{selection stability} via average pairwise overlap rate $|A\cap B|/|A|$ across runs.

\subsubsection{Results}

\begin{table*}[t]
\centering
\caption{Comparison between SCTA, Biomni, and the single-pass baseline across datasets.}
\label{tab:baseline_comparison}
\begin{tabularx}{\textwidth}{
l
p{6cm}
>{\raggedleft\arraybackslash}X
>{\raggedleft\arraybackslash}X
>{\raggedleft\arraybackslash}X
}
\toprule
\textbf{Dataset} & \textbf{Metric} & \textbf{SCTA} & \textbf{Biomni} & \textbf{Baseline} \\
\midrule
GSE193337 & Open Targets score (Q3)         & \textbf{0.12} & 0.04 & 0.12 \\
          & Enrichment Significant Pathways & \textbf{31}   & 0    & 0 \\
          & Mean Enrichment Score           & \textbf{306.90} & 68.99 & 183.75 \\
          & Average Runtime (seconds)       & 190.8 & \textbf{79.32} & 90.56 \\
          & Average Pairwise Overlap Rate   & \textbf{0.33} & 0.26 & -- \\
\midrule
GSE165045 & Open Targets score (Q3)         & \textbf{0.009} & 0.000 & 0.003 \\
          & Enrichment Significant Pathways & \textbf{94}    & 55   & 43 \\
          & Mean Enrichment Score           & 730.05 & \textbf{2984.04} & 529.49 \\
          & Average Runtime (seconds)       & 419.80 & 132.52 & \textbf{122.82} \\
          & Average Pairwise Overlap Rate   & \textbf{0.43} & 0.12 & -- \\
\midrule
GSE149689 & Open Targets score (Q3)         & 0.04 & 0.07 & \textbf{0.11} \\
          & Enrichment Significant Pathways & \textbf{81}   & 57   & 36 \\
          & Mean Enrichment Score           & 407.51 & 228.00 & \textbf{780.40} \\
          & Average Runtime (seconds)       & 318.74 & \textbf{160.16} & 172.17 \\
          & Average Pairwise Overlap Rate   & \textbf{0.60} & 0.44 & -- \\
\bottomrule
\end{tabularx}
\end{table*}

Table~\ref{tab:baseline_comparison} shows a stability-centered advantage for SCTA: it achieves stronger upper-tail disease association support (Q3) in two of three datasets, remains competitive in the remaining dataset, and consistently recovers more significantly enriched pathways across all datasets. Biomni attains lower runtimes but exhibits lower pairwise overlap rates, indicating that faster execution does not address the instability introduced by decision-dependent single-cell pipelines. SCTA's substantially higher overlap rates across independent runs demonstrate that decision-centric orchestration yields more reproducible target sets.

This difference matters operationally because translational target discovery is a low-throughput decision problem: downstream experiments typically validate only a small number of candidates. When target sets vary substantially across runs, the order in which genes are tested can change, altering both resource allocation and biological conclusions. Runtime therefore cannot be interpreted in isolation. A faster framework is less useful when it repeatedly shifts the shortlist that guides follow-up validation, whereas higher repeated-run overlap provides a more reliable basis for prioritizing costly experimental work.

\subsection{Ablation Analysis}
\label{sec:ablation_results}
\begin{figure*}[t]
\centering
\includegraphics[width=0.82\textwidth]{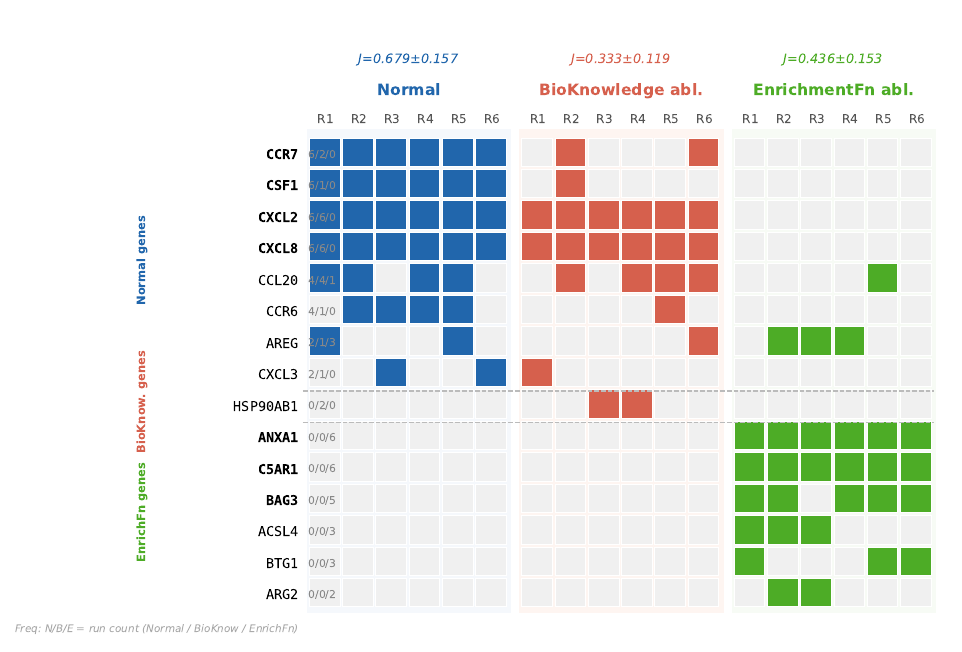}
\caption{Gene-selection matrix for hereditary CP under the fair six-run downstream-only ablation. Rows are genes recurring in at least two runs of any setting; columns are six runs per setting. Blue cells indicate Normal selection, orange indicates BioKnowledge ablation, green indicates EnrichmentFunction ablation, and gray indicates non-selection. Side frequency labels show run count per setting (Normal/BioKnow/EnrichFn). CCR7, CSF1, CXCL2, and CXCL8 appear in all six Normal runs; CCL20 and CCR6 in four of six.}
\label{fig:her_oncoprint}
\Description{Compact binary selection matrix with 15 gene rows and 18 run columns grouped into three settings. Blue = Normal, orange = BioKnowledge ablation, green = EnrichmentFunction ablation. CCR7, CSF1, CXCL2, CXCL8 selected in all six Normal runs.}
\end{figure*}

Table~\ref{tab:panc_her_ablation} summarizes the fair downstream-only ablation results for hereditary CP and Figure~\ref{fig:her_oncoprint} shows the compact six-run selection matrix. Under the fixed-artifact downstream protocol, the full SCTA configuration (Normal) achieved the highest within-setting stability, with mean pairwise Jaccard $0.679\pm0.157$. BioKnowledge\_ablation reduced stability most strongly ($0.333\pm0.119$), while EnrichmentFunction\_ablation retained intermediate stability ($0.436\pm0.153$) but converged to a distinct recurrent gene program.

\begin{table*}[t]
\centering
\caption{Fair downstream-only ablation for hereditary CP. All settings reuse fixed preprocessing artifacts; only the evidence module differs. Stability is measured by average pairwise Jaccard across six independent runs.}
\label{tab:panc_her_ablation}
\small
\begin{tabularx}{\textwidth}{l c X X}
\toprule
\textbf{Setting} & \textbf{Mean Jaccard $\pm$ Std.} & \textbf{Stable genes ($\geq$3/6)} & \textbf{Stable genes (6/6)} \\
\midrule
Normal & \textbf{0.679 $\pm$ 0.157} & CCL20, CCR6, CCR7, CSF1, CXCL2, CXCL8 & CCR7, CSF1, CXCL2, CXCL8 \\
BioKnowledge\_ablation & 0.333 $\pm$ 0.119 & CCL20, CXCL2, CXCL8 & CXCL2, CXCL8 \\
EnrichmentFunction\_ablation & 0.436 $\pm$ 0.153 & ACSL4, ANXA1, AREG, BAG3, BTG1, C5AR1 & ANXA1, C5AR1 \\
\bottomrule
\end{tabularx}
\end{table*}

The Normal setting preserved a recurrent inflammatory target core, with CCR7, CSF1, CXCL2, and CXCL8 selected in all six runs. CCL20 and CCR6 appeared in four of six Normal runs, supporting the relevance of the known CCR6--CCL20 hereditary CP axis without requiring that both genes persist in every run. In contrast, BioKnowledge\_ablation retained only CXCL2 and CXCL8 across all six runs and yielded the strongest numerical stability loss, indicating that network-context and concise biological-knowledge signals play an important stabilizing role in downstream prioritization.

EnrichmentFunction\_ablation was more stable than BioKnowledge\_ablation numerically, but its recurrent program shifted toward ANXA1 and C5AR1 as the only 6/6 genes, with ACSL4, AREG, BAG3, and BTG1 appearing at least 3/6 times. This pattern suggests that removing TF/pathway enrichment can still produce internally recurrent outputs while redirecting the final target program toward a different biological emphasis. Consistent with this interpretation, the Normal-stable genes are enriched for Inflammatory Response and TNF-alpha/NF-kB signaling in post-hoc pathway analysis, matching the inflammatory immune programs previously reported in hereditary CP.

Taken together, these results support a focused claim: under fixed preprocessing artifacts, SCTA's full evidence integration improves \emph{downstream target-selection stability} in hereditary CP relative to both ablated configurations. The comparison does not imply full end-to-end preprocessing stability, nor does it establish universal ablation superiority across diseases. Rather, it demonstrates that evidence-module design materially affects the reproducibility and biological coherence of final targets in this mechanistic case study.

\begin{figure*}[t]
\centering
\includegraphics[width=0.78\textwidth]{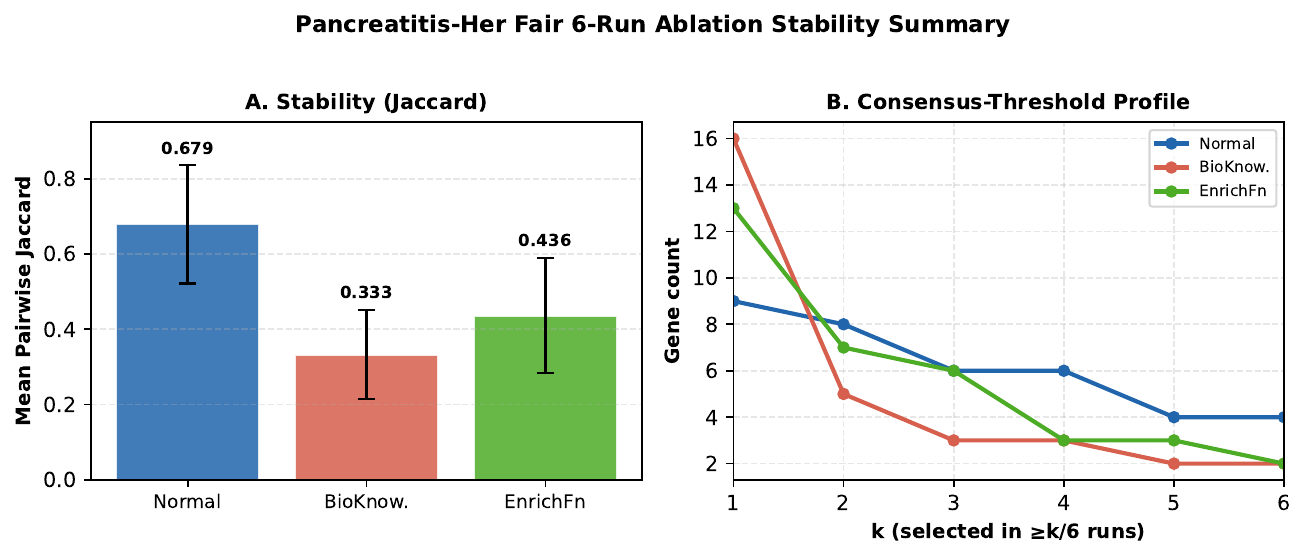}
\caption{Stability summary for the fair six-run downstream-only ablation. (A) Mean pairwise Jaccard similarity across settings. (B) Consensus-threshold profile: number of genes selected in at least $k/6$ runs as $k$ increases from 1 to 6.}
\label{fig:stability_summary}
\Description{Two-panel figure. Panel A shows a bar chart of mean pairwise Jaccard similarity for Normal (0.679), BioKnowledge ablation (0.333), and EnrichmentFunction ablation (0.436). Panel B shows consensus-threshold curves: Normal retains more genes at every threshold, with 4 genes at 6/6 versus 2 for each ablation.}
\end{figure*}

Figure~\ref{fig:stability_summary} provides a complementary view of these stability differences. Panel~A confirms that the full SCTA configuration achieves the highest mean pairwise Jaccard ($0.679\pm0.157$), approximately twice that of BioKnowledge\_ablation ($0.333\pm0.119$). Panel~B shows the consensus-threshold profile: as the stringency threshold $k$ increases from 1 to 6, Normal retains more genes at every level. At the strictest threshold ($k=6/6$), Normal preserves four genes (CCR7, CSF1, CXCL2, CXCL8), whereas both ablated settings retain only two. This indicates that SCTA's full evidence integration produces not only higher average overlap but also deeper consensus---a larger core of genes that the framework selects consistently regardless of run-to-run stochasticity.

Consensus depth complements average Jaccard because the two metrics capture different failure modes of stochastic prioritization. Mean pairwise Jaccard summarizes typical run-to-run agreement, but a configuration can achieve a moderately high pairwise score while still shuffling its top-$k$ shortlist across runs in ways that change which genes a downstream experimentalist would actually validate. Reporting that Normal retains four genes at the strictest $6/6$ threshold is therefore stronger evidence than mean overlap alone: it demonstrates that a non-trivial core of candidates survives every independent run, not merely that pairs of runs tend to agree on average. Read jointly with Table~\ref{tab:panc_her_ablation} and Figure~\ref{fig:her_oncoprint}, the two ablations expose distinct stability pathologies. BioKnowledge\_ablation collapses primarily to CXCL2 and CXCL8 at $6/6$, indicating that without network-context and concise functional evidence the framework loses the contextual grounding needed to consistently surface secondary but mechanistically related genes such as CCR7 and CSF1; the recurrent core becomes both numerically smaller and biologically narrower. EnrichmentFunction\_ablation, by contrast, remains internally recurrent at intermediate Jaccard but converges on a distinct program dominated by ANXA1 and C5AR1, illustrating that recurrence alone is insufficient: a setting can stably reproduce a target list that no longer reflects the inflammatory chemokine program independently characterized in hereditary CP. Together, these patterns support evaluating stability and biological coherence as complementary, not interchangeable, criteria.

\section{Case Studies}

\subsection{Agentic Identification of Etiology-Specific Immune Programs in Chronic Pancreatitis}

We applied SCTA to a published CITE-seq and scTCR-seq dataset of chronic pancreatitis (CP), profiling pancreatic immune cells from hereditary CP, idiopathic CP, and non-diseased controls. The original study reported an upregulated CCR6--CCL20 axis in hereditary CP, with enrichment of CCR6, CXCR4, GPR183, and CCL20 in CD4$^+$ and CD8$^+$ T cells \cite{lee2022single}.

\subsubsection{Hereditary CP: Reproducing the CCR6--CCL20 Axis and Beyond}
SCTA identified both CCR6 and CCL20 among its prioritized genes in T cells, recapitulating the core ligand--receptor axis validated in the original study without explicit supervision. Under the fair downstream-only ablation protocol, both genes remained part of the recurrent Normal-setting program, with CCL20 and CCR6 appearing in four of six runs, while CCR7, CSF1, CXCL2, and CXCL8 formed a stricter 6/6 stable core. This links SCTA's stability signal to a known hereditary CP inflammatory mechanism rather than to a purely numerical recurrence artifact.

The CCR6--CCL20 axis plays a central role in lymphoid homing and canonical immune trafficking \cite{hauser2016common}. CCR6 is a chemokine receptor associated with lymphoid tissue organization, and CCL20 is its cognate ligand. The upregulation of this axis in hereditary CP reflects a shift toward immune-driven inflammatory remodeling in pancreatic tissue. SCTA's ability to recover both components without explicit supervision demonstrates that the framework can identify biologically coherent ligand--receptor pairs through independent evidence aggregation across cell types and functional annotations.

The stricter 6/6 Normal core further sharpens this interpretation by linking recurrence to multiple, complementary immune mechanisms rather than to a single dominant signal. CCR7 is a canonical trafficking receptor involved in lymphocyte migration and immune-cell positioning, while CSF1 supports myeloid survival and macrophage lineage signaling; together they suggest stable recovery of coordinated lymphoid--myeloid communication rather than isolated marker expression. CXCL2 and CXCL8 add a second layer centered on inflammatory chemokine activity and neutrophil recruitment, indicating that the recurrent core captures both immune-cell organization and active inflammatory effector programs. Against this background, the 4/6 recurrence of CCL20 and CCR6 is also informative: these genes are not merely occasional outputs, but a repeatedly recovered hereditary-CP-specific axis consistent with the published disease mechanism. Taken together, the hereditary CP results show that SCTA recovers known biology while also prioritizing additional interpretable candidates that extend the mechanistic picture without requiring claims of experimental validation.

\begin{figure}[ht]
\centering
\includegraphics[width=0.95\columnwidth]{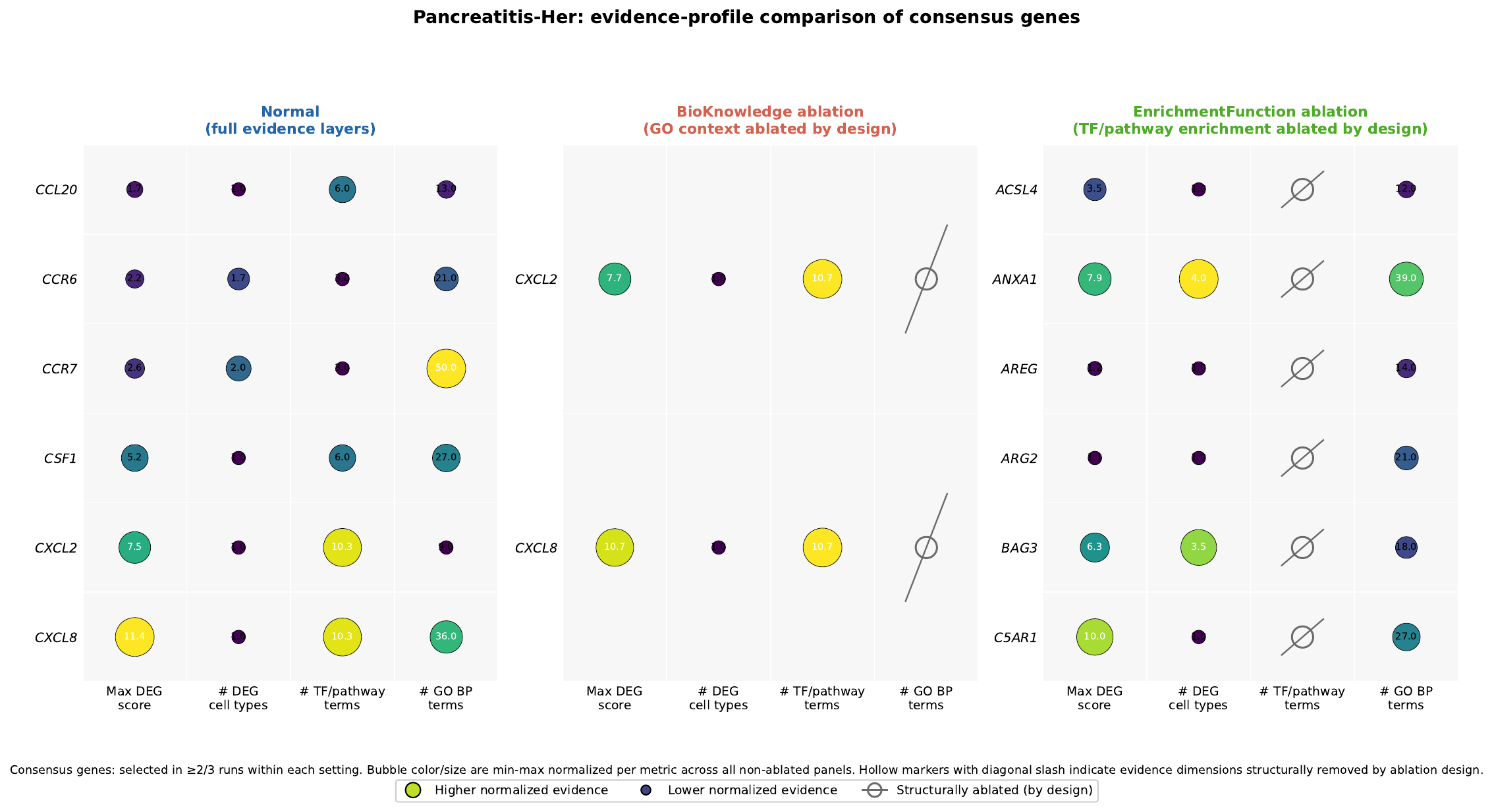}
\caption{Evidence profiles for representative recurrent genes across ablation settings in hereditary CP. Three panels compare the evidence dimensions retained by the fair downstream-only ablation protocol: maximum DEG score, number of DEG-supported cell types, number of TF/pathway terms, and number of GO biological-process terms. Hollow slashed markers indicate evidence dimensions structurally removed by the ablation design. The full SCTA configuration preserves a coherent inflammatory recurrent core, whereas ablated settings either contract the recurrent gene set or shift it toward a different program.}
\label{fig:her_evidence}
\Description{A three-panel figure comparing evidence profiles across ablation settings for hereditary chronic pancreatitis. The left panel (Normal) shows recurrent genes under full evidence integration. The middle panel (BioKnowledge ablation) shows recurrent genes identified without network-context evidence. The right panel (EnrichmentFunction ablation) shows recurrent genes identified without TF/pathway enrichment. Each gene is represented by four evidence dimensions: maximum DEG score, number of DEG-supported cell types, number of TF/pathway terms, and number of GO biological-process terms.}
\end{figure}

Beyond CCR6 and CCL20, SCTA highlighted multiple chemokines and inflammatory mediators with cell-type-specific attribution. CXCL2 was prioritized in macrophages and CXCL8 (IL-8) in monocytes, both enriched in cytokine--cytokine receptor interaction and TNF signaling pathways. CXCL2 and CXCL8 are ELR$^+$ CXC chemokines that drive neutrophil recruitment via CXCR2/CXCR1, and this chemotactic axis contributes to inflammatory infiltration and acinar injury during acute and chronic pancreatitis \cite{russo2014cxcl8,hertzer2013cxcr2}. In mouse models, genetic deletion or pharmacologic inhibition of CXCR2 reduces neutrophil recruitment and protects against pancreatic inflammation \cite{steele2015cxcr2}. In human chronic pancreatitis tissues, IL-8 is significantly enriched in immune cells surrounding enlarged pancreatic nerves, with additional staining in metaplastic ductal cells, supporting a role for CXCL8 in local inflammatory remodeling \cite{di2000expression}. That both CXCL2 and CXCL8 appear in the 6/6 stable Normal core reinforces the biological coherence of the recurrent target set: these genes converge on the same neutrophil-recruitment pathway through independent cell-type attributions.

Amphiregulin (AREG), a membrane-bound EGFR ligand involved in epithelial repair programs, was also prioritized in T cells. Previous studies have associated AREG expression with CP in pre-clinical models, where it modulates the balance between tissue repair and fibrotic remodeling \cite{glaubitz2022mouse}. SCTA further identified stress-associated signals linked to CCR6$^+$ T cells, including HSPA8, HSP90AA1, DNAJB1, and PPP1R15A. This finding is consistent with previous CP studies reporting chronic activation of endoplasmic reticulum stress and unfolded protein response pathways as a key feature of pancreatic inflammation \cite{sah2014endoplasmic}. Taken together, the targets proposed by SCTA complement the original study by offering additional cell-type-resolved perspectives on drug target discovery, highlighting genes with potential as novel therapeutic targets for hereditary CP.

\subsubsection{Idiopathic CP: Distinct Immune Signals}
Idiopathic CP (ICP) is a fibroinflammatory pancreatic disease in which patients develop progressive pancreatic injury, chronic pain, and exocrine/endocrine dysfunction without an identifiable cause \cite{cartelle2023natural}. Clinically, ICP represents an unmet medical need: current management is largely supportive and does not directly halt the underlying inflammatory--fibrotic progression that drives irreversible organ damage \cite{hobbsmanagement}.

As expected, CCR6 was absent from the SCTA output for ICP, consistent with the original study's observation that idiopathic CP lacks the CCR6-driven immune response seen in hereditary disease \cite{hauser2016common}. Instead, immune signals were dominated by CCR7 in B cells, a chemokine receptor associated with lymphoid homing and canonical immune trafficking \cite{hauser2016common}. This shift mirrors the etiology-specific immune program distinction reported in the original dataset: hereditary CP is characterized by CCR6--CCL20-mediated T cell recruitment, whereas ICP exhibits a different immune trafficking signature.

JUNB, DUSP1, and HSPH1 emerged as prioritized genes from the agentic workflow analysis of the idiopathic CP dataset. Collectively, these genes are linked to stress-activated and MAPK-associated signaling programs, consistent with prior literature on inflammatory activation and cellular stress adaptation \cite{leppa1999diverse,boutros2008map,vydra201917,gurzov2008junb}. While these pathways are pharmacologically druggable, translating MAPK-pathway modulation to ICP will likely require cell-type targeted delivery and careful dosing to mitigate systemic on-target toxicities. These targets should therefore be interpreted as computational hypotheses for further experimental investigation rather than validated therapeutic candidates.

This case study demonstrates that SCTA can (i) reproduce the central CCR6--CCL20 axis specific to hereditary CP while extending the target space, (ii) provide mechanistic interpretations grounded in cellular context, and (iii) uncover distinct, etiology-specific therapeutic opportunities in idiopathic CP that reflect the underlying biological differences between disease subtypes.

\section{Discussion}
The fair downstream-only ablation protocol is scientifically useful because it isolates evidence-module effects from upstream preprocessing variability. In the earlier mixed-protocol comparison, the Normal setting included fresh preprocessing while the ablated settings reused fixed artifacts, making it impossible to attribute stability differences cleanly to evidence design. By holding preprocessing artifacts fixed and re-running only the downstream target-selection stages, the updated hereditary CP experiment focuses on the specific question of whether SCTA's biological evidence modules improve the reproducibility of final target prioritization.

Under this protocol, the full SCTA configuration preserved a coherent inflammatory recurrent core centered on CCR7, CSF1, CXCL2, and CXCL8, with CCL20 and CCR6 recurring in four of six runs. This pattern is consistent with immune recruitment and inflammatory signaling processes previously reported in hereditary CP, and the post-hoc enrichment signal for Inflammatory Response and TNF-alpha/NF-kB signaling reinforces that the recurrent Normal targets are not only numerically stable but biologically coherent. In contrast, EnrichmentFunction\_ablation retained moderate numerical stability while shifting toward a distinct recurrent program dominated by ANXA1 and C5AR1, suggesting that a setting can remain internally recurrent yet become biologically reoriented when pathway-level evidence is removed.

These findings also illustrate a practical runtime--stability tradeoff. SCTA is slower than simpler baselines and broad automation systems because it explicitly gathers and integrates multiple evidence streams, but this added structure yields more reproducible final target sets and clearer mechanistic interpretation. For translational discovery, such additional computation may be justified when candidate prioritization rather than exploratory screening is the primary objective.

More broadly, repeated-run consensus offers a practical reliability check for any stochastic agent workflow. LLM-based agents can produce different outputs across runs due to sampling variability, prompt sensitivity, and non-deterministic tool execution. When the downstream cost of acting on a wrong candidate is high---as in drug target validation---running the framework multiple times and retaining only recurrent selections provides a simple but effective filter against run-specific artifacts. However, stable recurrence should not be treated as sufficient validation by itself: a framework could converge on a numerically stable but biologically irrelevant gene set. The hereditary CP case study illustrates the complementary role of external biological evidence in grounding the stability signal, and we recommend that repeated-run consensus always be interpreted alongside independent mechanistic support.

The chronic pancreatitis case study is particularly informative because it connects numerical stability to independently validated disease biology. Recurrence alone is insufficient evidence of biological relevance---a framework could converge on a stable but biologically arbitrary gene set. In hereditary CP, the recurrent Normal core (CCR7, CSF1, CXCL2, CXCL8) and the 4/6-recurrent CCR6--CCL20 axis correspond to inflammatory chemokine and immune trafficking programs independently characterized in prior mechanistic studies \cite{lee2022single,russo2014cxcl8,hertzer2013cxcr2}, providing external grounding for the stability signal. The etiology-specific contrast between hereditary and idiopathic CP further demonstrates that SCTA's outputs reflect genuine biological differences rather than a generic inflammatory signature.

These observations carry practical implications for how agentic biomedical workflows should be deployed and evaluated. In translational settings, repeated-run consensus can serve as a lightweight but informative filter: candidates that recur across independent runs are less likely to reflect prompt-specific or sampling-specific artifacts, and more likely to represent signals that the evidence base consistently supports. At the same time, stability must always be interpreted alongside biological evidence, because a framework can converge on a numerically stable but mechanistically uninformative gene set---as illustrated by the EnrichmentFunction\_ablation shift toward ANXA1 and C5AR1. The fair downstream-only protocol used in this study is particularly useful for isolating evidence-module effects from upstream preprocessing variability, but it does not capture the full spectrum of perturbations that arise in end-to-end single-cell analysis. Future work should evaluate stability under broader perturbation regimes, including variation in normalization, integration, and clustering parameters. In all cases, SCTA is best understood as a prioritization framework that narrows the candidate space for experimental follow-up, not as a replacement for wet-lab validation of therapeutic hypotheses.

Several limitations remain. First, hereditary CP is a focused mechanistic case study rather than a claim of universal cross-dataset ablation behavior. Second, the present ablation isolates downstream target-selection stability under fixed preprocessing artifacts; it does not establish full end-to-end stability of normalization, clustering, annotation, or differential-expression stages. Third, disease-specific effects likely matter: the stabilizing role of a given evidence module may differ across inflammatory, oncological, or infectious contexts. Finally, SCTA currently produces computational hypotheses only, and wet-lab validation remains necessary before therapeutic conclusions can be drawn.

\section{Conclusions}
We introduced SCTA, a decision-centric multi-agent framework for target gene discovery from scRNA-seq data. By decomposing analysis into specialized agents and explicitly modeling decision-critical stages, SCTA improves the stability and interpretability of final target prioritization relative to single-pass baselines and broad automation frameworks.

Across the evaluated datasets, SCTA provides competitive disease association support and strong pathway-level coherence while improving repeated-run target convergence. In hereditary chronic pancreatitis, the fair downstream-only six-run ablation shows that SCTA's full evidence integration improves downstream target-selection stability under fixed preprocessing artifacts and preserves a coherent inflammatory recurrent core. These findings support SCTA as a practical framework for stable and interpretable target discovery, while leaving full end-to-end stability analysis, broader disease-specific ablation studies, and wet-lab validation to future work.

\bibliographystyle{ACM-Reference-Format}
\bibliography{references}

\end{document}